\documentclass[]{bytedance_seed}

\usepackage[toc,page,header]{appendix}

\usepackage{minitoc}

\usepackage{amssymb}

\usepackage{subcaption}
\usepackage{xcolor}
\usepackage{xspace}
\usepackage{varwidth}
\usepackage{pifont} 
\usepackage{wrapfig}
\usepackage[ruled,vlined]{algorithm2e}
\usepackage{CJKutf8}
\usepackage{pgfplots}
\pgfplotsset{compat=1.18}
\usepackage{enumitem}

\title{
Protenix-Mini: Efficient Structure Predictor via Compact Architecture, Few-Step Diffusion and Switchable pLM
}

\author[1]{Chengyue Gong}
\author[1]{Xinshi Chen}
\author[1]{Yuxuan Zhang}
\author[2]{Yuxuan Song}
\author[2]{Hao Zhou}
\author[1]{Wenzhi Xiao}

\affiliation[1]{ByteDance Seed}
\affiliation[2]{Tsinghua University}

\newcommand{\xs}[1]{{\color{red}{[xs: #1]}}}

\abstract{
Lightweight inference is critical for biomolecular structure prediction and other downstream tasks, enabling efficient real-world deployment and inference-time scaling for large-scale applications. 
In this work, we address the challenge of balancing model efficiency and prediction accuracy by making several key modifications: 1) Multi-step AF3 sampler is replaced by a few-step ODE sampler, significantly reducing computational overhead for the diffusion module part during inference; 2) In the open-source Protenix framework, a subset of pairformer/diffusion transformer blocks doesn't make contributions to the final structure prediction, presenting opportunities for architectural pruning and lightweight redesign;
3)  A model incorporating an ESM module is trained to substitute the conventional MSA module, reducing MSA preprocessing time.

Building on these key insights, we present \textbf{Protenix-Mini}, a compact and optimized model designed for efficient protein structure prediction. 
This streamlined version incorporates a more efficient architectural design with a two-step Ordinary Differential Equation (ODE) sampling strategy. 
By eliminating redundant Transformer components and refining the sampling process, Protenix-Mini significantly reduces model complexity with slight accuracy drop. 
Evaluations on benchmark datasets demonstrate that it achieves high-fidelity predictions, with only a negligible 1–5\% decrease in performance on benchmark datasets compared to its full-scale counterpart. 
This makes Protenix-Mini an ideal choice for applications where computational resources are limited but accurate structure prediction remains crucial.

}

\date{\today}
\begin{document}
\maketitle

\newcommand{\figref}[1]{Figure~\ref{#1}}
\newcommand{\todo}[1]{\colorbox{yellow!30}{[{#1}]}}

\newcommand{\outline}[1]{%
  \colorbox{cyan!20}{%
    \begin{varwidth}{\dimexpr\linewidth-2\fboxsep\relax}
    [{\bf Outline}]  #1
    \end{varwidth}%
  }%
}

\section{Introduction}
Accurate prediction of biomolecular structures, such as proteins and nucleic acids, is the cornerstone of modern structural biology and drug discovery, allowing insight into molecular function, interactions, and rational design of therapeutic agents \citep[e.g.][]{jumper2021highly,wang2023scientific,cheng2023accurate}.
However, state-of-the-art structure predictors (e.g. AlphaFold3 \citep{abramson2024accurate}, Chai-1 \citep{chai2024chai}, Protenix \citep{bytedance2025protenix}, Boltz-1 \citep{wohlwend2024boltz}), while achieving remarkable accuracy in many cases,
often rely on deep neural architectures and computationally intensive multiple step diffusion sampling procedures.
This poses significant challenges in resource-constrained environments (e.g., high-throughput screening pipelines) or large-scale inference tasks \citep{cao2022design}. Lightweight inference frameworks that balance efficiency and accuracy are therefore critical to bridging the gap between cutting-edge research and practical applications in biomedicine, computational protein design \citep{watson2023novo, pacesa2024bindcraft}, and inference-time scaling \citep{ma2025inference, liu2023flowgrad}.
In this study, we identify several orthogonal pathways to achieve faster inference:

\ding{182} \textbf{Efficient Sampling via Few-Step ODE.}
Recent advances in score-based generative modeling have shown that the diffusion process can be approximated using ordinary differential equations (ODEs), enabling deterministic sampling that bypasses the stochasticity of traditional SDE solvers \citep[e.g.][]{lu2022dpm, yang2023diffusion, liu2023instaflow, song2020denoising}. 
Our analysis reveals that models trained with
different generative model frameworks (e.g., EDM \citep{karras2022elucidating} in AlphaFold3 and flow matching \citep{liuflow, liu2022rectified, lipman2022flow})
both exhibit surprising robustness to drastic reductions in sampling steps. 
Specifically, we observe that these models can generate structurally accurate conformations even with as few as 2 ODE steps in practice, challenging the conventional wisdom that high-fidelity sampling requires tens of iterations. 
% Beyond enhancing sampling efficiency, this work also opens up new research frontiers and methodological insights in noise space optimization, structural editing, and architectural design.

\begin{figure*}[t]
\setlength{\tabcolsep}{8pt}
\centering
\begin{tabular}{cc}
\includegraphics[width=0.64\textwidth]{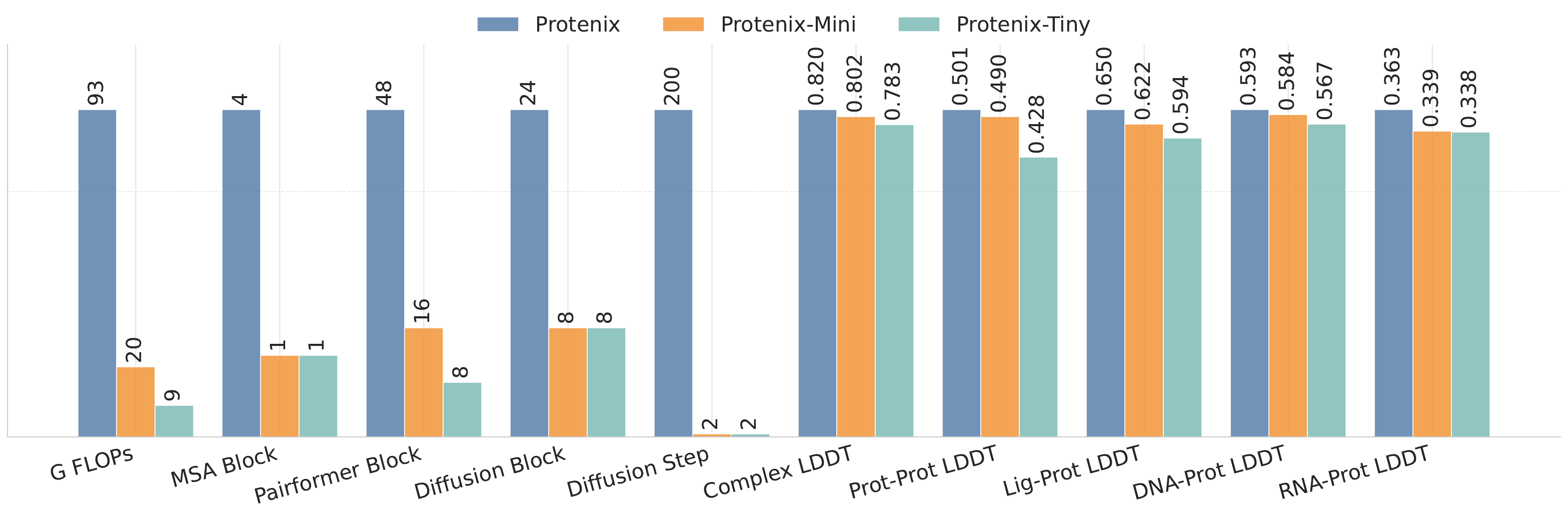}
&
\hspace{-20pt}
\includegraphics[width=0.35\textwidth]{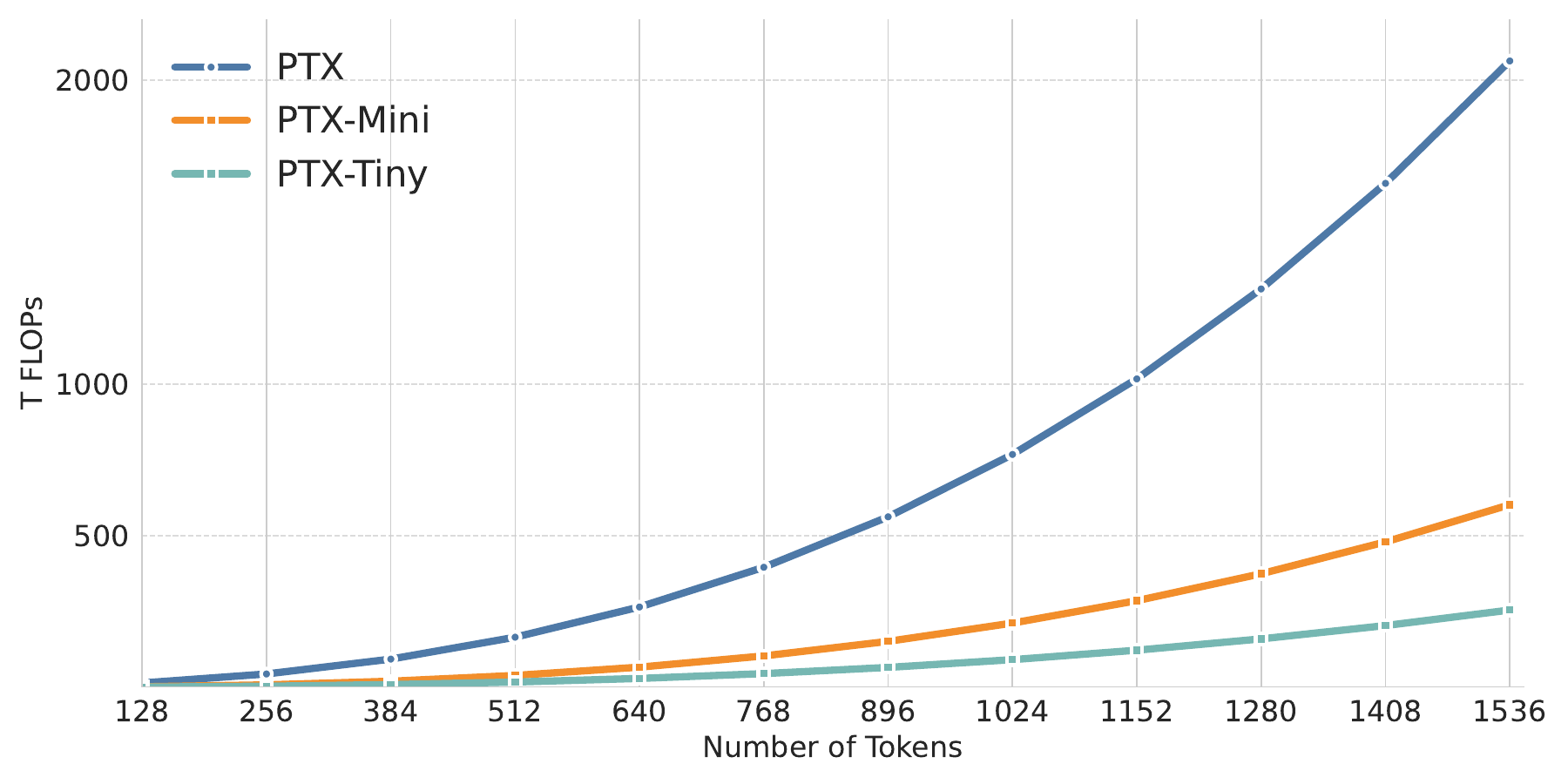} \\

\end{tabular}
\caption{\textbf{We elucidate the key configurations balancing model computation cost and performance on RecentPDB dataset.} 
In the right panel,
for efficiency optimization, the Mini and Tiny variants both reduce the number of MSA module blocks and pairformer blocks while truncating diffusion steps to two. Notably, these designs result in marginally lower performance compared to the standard Protenix model, with trade-offs carefully calibrated to maintain functional accuracy. All the numbers in the paper are evaluated on proteins with fewer than 768 tokens in the test set. We list more details in Table \ref{tab:appendix:architecture-config}.
% \textbf{FLOPs comparison of different models under varying token counts and MSA sizes.} 
In the right panel, a comparison of FLOPs across different models under varying token counts and MSA sizes are displayed. the number of MSAs is fixed at 2048, while the right panel sets the token length to 384. The number of atoms is fixed at 8832. 
% Notably, both the Mini and Tiny model variants exhibit significantly lower computational costs compared to the standard Protenix model, highlighting their efficiency advantages. \cy{update the right figure}
% in resource-constrained scenarios.
}
\label{fig:flops}
\end{figure*}

\ding{183} \textbf{Compression of Redundant Components.}
Analyzing the open source Protenix framework--an open source diffusion-based structure predictor, we discover that a subset of its pairformer/diffusion transformer blocks contribute negligibly to the final structure prediction. 
During post-training, directly removing these blocks during inference yields slightly worse results. 
By identifying and removing such noncritical components, we reconfigure the model architecture  to improve the inference efficiency.
As demonstrated in Figure \ref{fig:flops}, we arrive at much smaller FLOPs cost for different numbers of tokens and MSAs.

\ding{184} \textbf{Replace MSA Module with pLM.}
Given the high computational costs associated with both MSA search and MSA computation modules, we propose substituting the MSA component with embeddings extracted from the pretrained protein language model (pLM) to mitigate computational overhead. In practice, we use the ESM2-3B model \citep{lin2023evolutionary} to get the representations.

% \begin{figure}[t]
%     \centering
%     \includegraphics[width=1.0\linewidth]{figs/performance_comparison_flops.pdf}
%     \caption{
%     \textbf{We elucidate the key configurations balancing model computation cost and performance on RecentPDB dataset.} For efficiency optimization, the Mini and Tiny variants both reduce the number of MSA module blocks and pairformer blocks while truncating diffusion steps to two. Notably, these designs result in marginally lower performance compared to the standard Protenix model, with trade-offs carefully calibrated to maintain functional accuracy. All the numbers in the paper is evaluated on proteins with fewer than 768 tokens in the test set. \cy{tiny-performance to be updated, further train with large crop size}
%     % \cy{flops number (20 is small, 65 is 2-step large, tiny is 10) architecture configuration and numbers is wrong, change later}
%     }
%     \label{fig:flops}
% \end{figure}

Building on these insights, we introduce \textbf{Protenix-Mini}, a lightweight variant of Protenix that employs a reduced number of blocks and a few ODE steps (e.g., one or two) to enable efficient prediction of complex biomolecular structures. Experimental results show that Protenix-Mini achieves a favorable balance between efficiency and accuracy, with only a marginal 1–5\% reduction in evaluation metrics such as interface LDDT, complex LDDT, and ligand RMSD success rate. We demonstrate that lightweight models can rival their heavyweight counterparts in biomolecular structure prediction.
By addressing both sampling efficiency and architectural overhead, Protenix-Mini represents a step toward democratizing access to high-fidelity biomolecular structure prediction.
Looking ahead, we aim to further improve inference efficiency through several complementary approaches, e.g., architecture design, distillation, quantization, etc. 
% We plan to explore advanced attention mechanisms—such as sparse or dynamic attention—to better optimize the transformer backbone, particularly for complexes with a large number of tokens. We also aim to train improved protein language models to enhance prediction accuracy at molecular interfaces.

\section{Preliminary: Inference Procedure}

\begin{figure*}[b]
    \centering
    \includegraphics[width=1\linewidth]{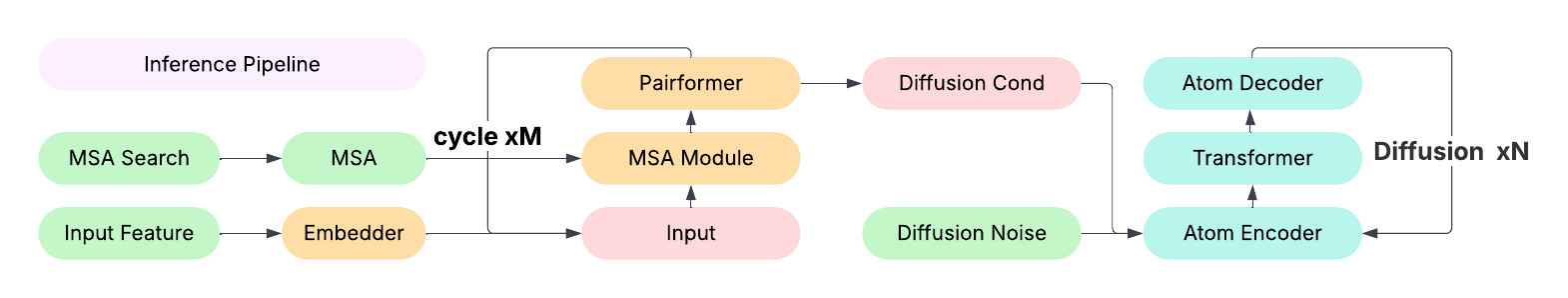}
    \vspace{-20pt}
    \caption{\textbf{Overview of the AF3-style model architecture.} The model consists of a conditioning stage that produces latent representations (\texttt{Diffusion Cond}) from input features, and a diffusion stage that transforms random initialization (\texttt{Diffusion Noise}) into the final atomic structure.
    }
    \label{fig:architecture}
\end{figure*}

This section provides a brief overview of the architecture and inference procedure shared by AF3-style structure prediction models~\citep[e.g.,][]{abramson2024accurate,bytedance2025protenix,wohlwend2024boltz,chai2024chai}. The overall inference pipeline, depicted in Figure~\ref{fig:architecture}, comprises two core stages: conditioning and diffusion.

\textbf{1. Conditioning.} The model first computes conditioning signals. Let $\texttt{Seq}$ denote the input sequence and $\texttt{MSA}$ the multiple sequence alignment. The MSA results and sequence embeddings are passed through the MSA module and Pairformer for $M$ cycles, yielding per-token and pairwise representations, denoted as $s$ and $z$ respectively:
$
(s, z) = \mathrm{Cond}(\texttt{Seq}, \texttt{MSA}).
$
These serve as conditioning inputs for the subsequent diffusion module.

\textbf{2. Diffusion.}  The diffusion process generates the 3D atomic structure by progressively denoising an initial random sample. Let $x_1$ denote the ground-truth structure, and $x_0 \sim \mathcal{N}(0, \sigma_\mathrm{data}^2 \mathbf{I})$ be the initial Gaussian noise. 
As in standard diffusion models, structure generation follows a forward–reverse paradigm:
\[
x_1 \xrightarrow{\text{forward}} x_0, \quad x_0 \xrightarrow{\text{reverse}} \hat{x}_1 \approx x_1,
\]
where the forward process perturbs the structure by progressively adding noise, and the reverse process attempts to recover the original signal by applying a learned denoiser. In AF3, this is implemented through a sampling algorithm, summarized in Algorithm~\ref{alg:edmedm-sample}, following the EDM formulation \citep{karras2022elucidating}. 
At each iteration, noise $\epsilon_t$ is injected into the current sample $x_t$ according to a predefined schedule, and the Diffuser predicts a denoised version from the perturbed input. A velocity is then computed and applied to update $x_t$, scaled by a step-size parameter $\eta$. 
%While the original motivation for using $\eta > 1$ is not explicitly stated, prior work—such as Auto Guidance~\citep{karras2024guiding}—suggests that denoisers tend to underestimate velocity magnitudes in low-noise regimes ($t \rightarrow 1$), leading to oversmoothed outputs. Although we have not directly verified this effect in our setting, we find that a larger $\eta$ empirically improves the quality of the generated structures. 
Additional implementation details are provided in the supplementary materials of~\citep{abramson2024accurate}.

% At each continuous time $t \in [0, 1]$, the model predicts the denoised structure:
% \[
% x_t^\mathrm{denoised} \leftarrow \mathrm{Diffuser}(x_t, t \mid s, z, \texttt{Seq}, \texttt{MSA}).
% \]
% This predicted $x_t^\mathrm{denoised}$ is then used to compute a velocity update or reconstruction target.
% After iterating over $t \in [0, 1]$, the final structure $\hat{x}_1$ is obtained.

%Notably, the sampling procedure is decoupled from training and can be adapted to an ODE-based sampler. In this variant, summarized in Algorithm~\ref{alg:rectflow-sample} (Appendix), the Diffuser operates directly on $x_t$ without added noise, and the step size is fixed to $\eta = 1$. We hypothesize this configuration is most effective in the low-noise regime ($t \rightarrow 1$). In the following section, we compare the impact of these different sampling strategies.

\begin{algorithm}[H]
\caption{\textbf{Sampling Algorithm in AF3}} 
\label{alg:edmedm-sample}
\textbf{Given:} $\eta = 1.5$, $\gamma_0 = 0.8$, $\gamma_{\min} = 1$, $\lambda = 1.003$, $\sigma_{\text{data}} = 16$, and Diffuser with condition $C$.

\textbf{Initial:} $x_0 \sim \mathcal{N}(0, \sigma_{\text{data}} \mathbf{I}), \; t = 0$

$\texttt{StepScheduler}=\mathrm{StepSchedule}(\sigma_{\text{data}})$

\While{$t < 1$}{
    $\Delta t \leftarrow \mathrm{next(\texttt{StepScheduler})}$

    $x_t \leftarrow \mathrm{CenterRandomAugmentation}(x_t)$ \tcp*[r]{centered and randomly rotated and translated}
    
    $\epsilon_t, \hat{t} \leftarrow \mathrm{NoiseSchedule}\bigg (\Delta t, t, \gamma_0, \gamma_{\min}, \mathcal{N}(0, \mathbf{I}) \bigg )$ \tcp*[r]{time-dependent noise scale}
    
    $x_t^\mathrm{noisy} \leftarrow x_t + \lambda \epsilon_t$ \tcp*[r]{add scaled noise}
    
    $x_t^\mathrm{denoised} \leftarrow \mathrm{Diffuser}(x_t^\mathrm{noisy}, \hat{t} \mid C)$ \tcp*[r]{predict denoised structure}

    $t \leftarrow t + \Delta t$ \tcp*[r]{advance time}

    $x_t \leftarrow x_t^\mathrm{noisy} + \eta \cdot (t-\hat{t}) \cdot \mathrm{CalVelocity}(x_t^\mathrm{denoised}, x_t^\mathrm{noisy}, \hat{t})$ \tcp*[r]{update via learned velocity}
}
\end{algorithm}

\section{Lightweight Structure Predictor}

% \subsection{Reducing Diffusion Steps}
\subsection{Enabling Few-Step Diffusion via Sampler Configuration} 

One of the primary bottlenecks in diffusion-based structure generation is the high computational cost incurred by long sampling trajectories. AF3-style models default to samplers with up to 200 steps~\citep{abramson2024accurate}. But do such long sampling schedules really need to be preserved at inference time? Surprisingly, the answer is no—\textbf{as long as the sampling algorithm is appropriately configured}. We find that AF3-style models remain highly effective even under drastically reduced sampling steps—\textbf{without any retraining}. For instance, 200-step sampling can be replaced with as few as 2 ODE steps while still achieving strong performance. Even one-step inference yields reasonable predictions with only modest degradation. 

When directly applying the default AF3 sampler with fewer inference steps, we found that performance collapsed below 10 steps—often 
producing broken or nonsensical structures. Upon further analysis, we found that this degradation was not caused by limitations in the model itself, but rather by the default sampling algorithm, which was poorly suited for few-step inference.  Through a systematic analysis of the sampling procedure, we identified two simple yet crucial changes that unlock effective few-step inference:
\begin{itemize}[leftmargin=*]
\item switching to an \emph{ODE sampler} by setting \emph{$\gamma_0 = 0$}, thereby removing added noise; 
\item setting the \emph{step scale $\eta = 1.0$}, which restores consistency with the underlying velocity-based formulation.
\end{itemize}

\begin{figure}[h!]
  \centering
  \begin{subfigure}[t]{0.4\textwidth}
    \centering
    \includegraphics[width=\linewidth]{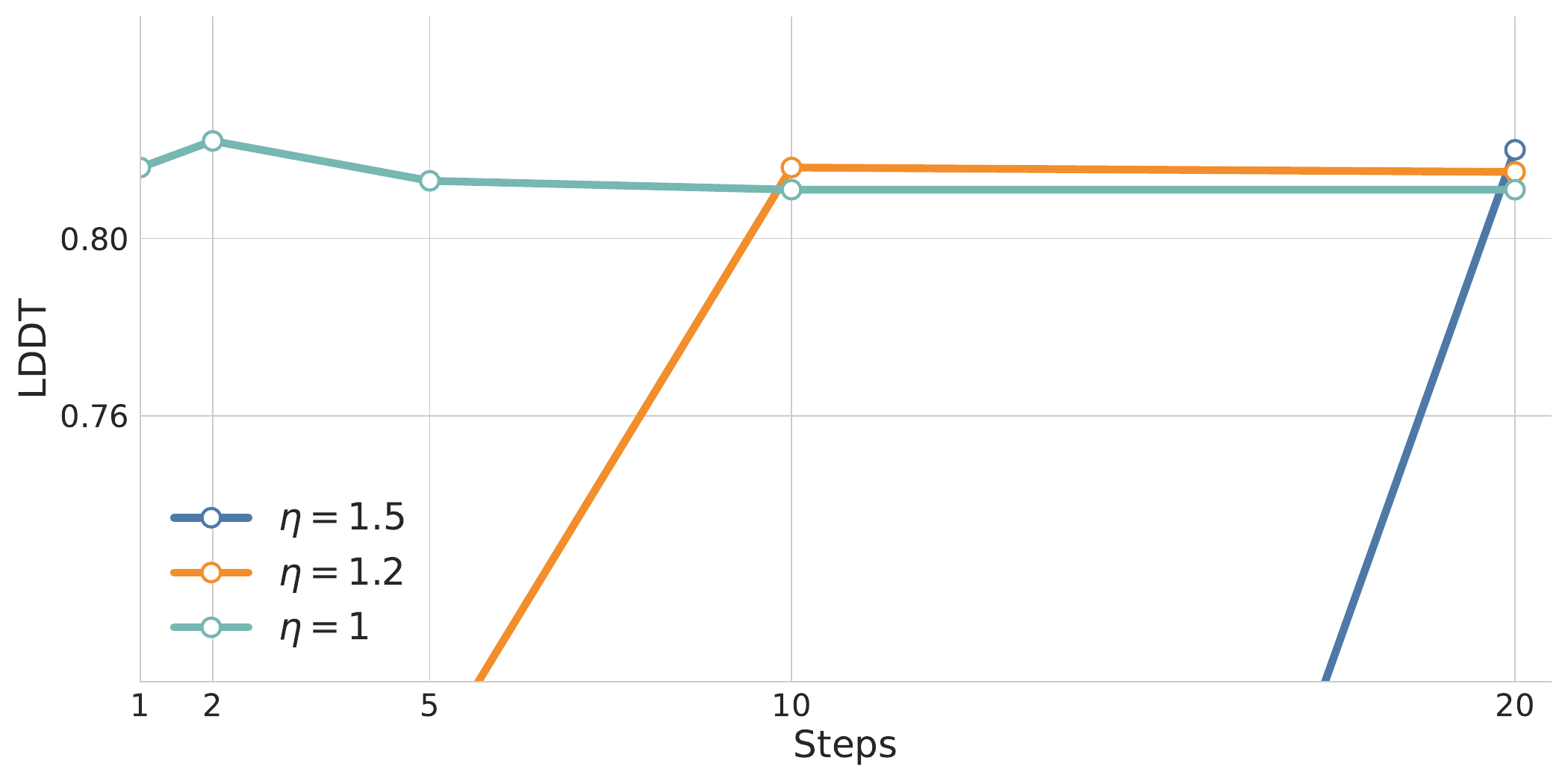}
    \caption{LDDT score across different $\eta$ and steps.}
  \end{subfigure}
  \hfill
  \begin{subfigure}[t]{0.4\textwidth}
    \centering
    \includegraphics[width=\linewidth]{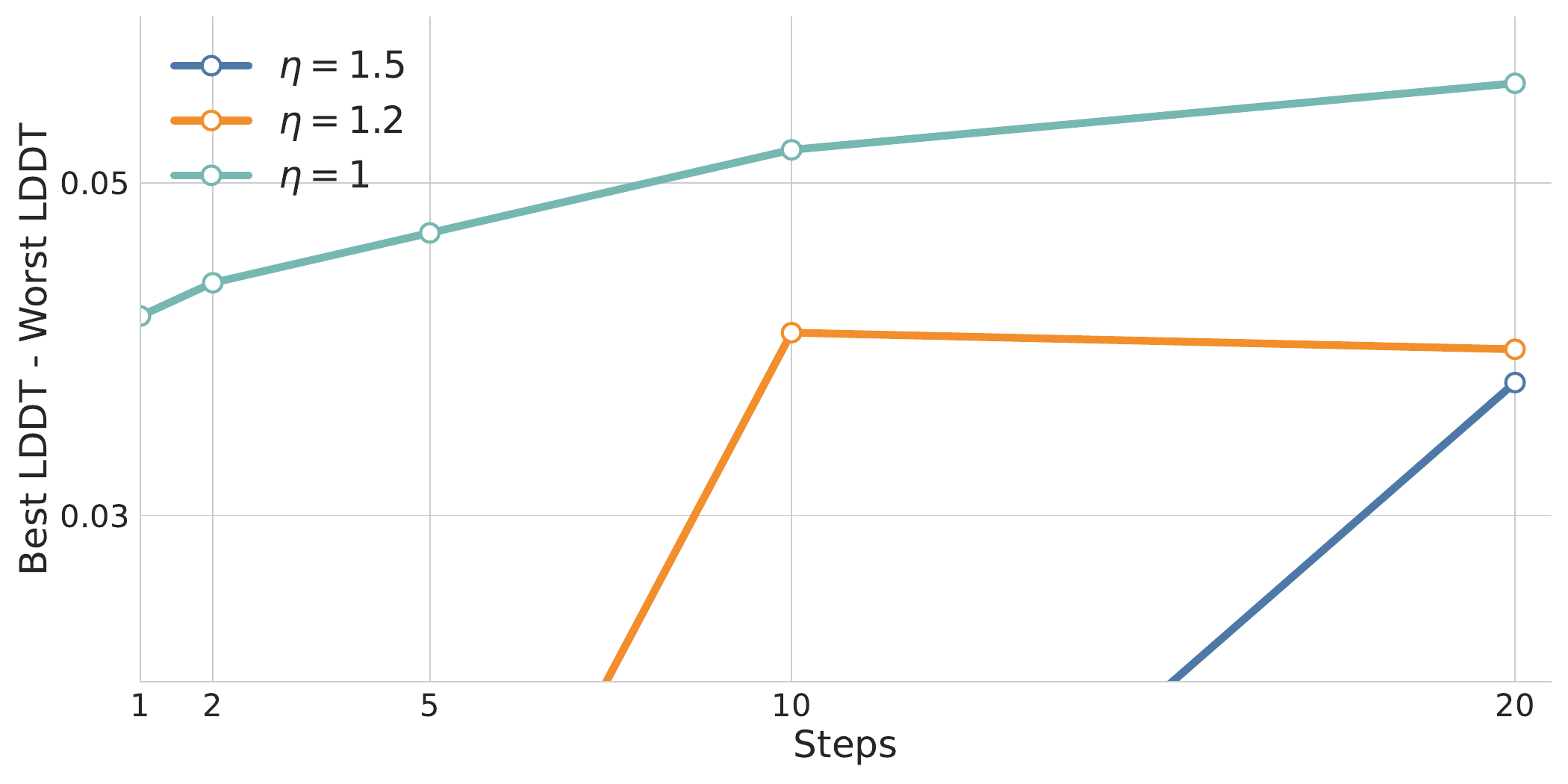}
    \caption{LDDT variance across $\eta$ and steps.}
  \end{subfigure}
  \caption{
    \textbf{Impact of step scale $\eta$ and number of steps.}
    Performance variation of an ODE sampler across different settings, showing trends in accuracy and variance. The y-axis shows the complex LDDT scores, and the x-axis displays the number of diffusion steps during inference.
  }
  \label{fig:eta-study-bar}
\end{figure}

A central factor affecting few-step performance is the step scale $\eta$, which was introduced in AF3 as a modification to the EDM sampler~\citep{karras2022elucidating}. EDM uses $\eta = 1$ by default, and AF3 increases this to $\eta = 1.5$. While the original motivation for using $\eta > 1$ is not explicitly stated, prior work—such as Auto Guidance~\citep{karras2024guiding}—suggests that denoisers tend to underestimate velocity magnitudes in low-noise regimes ($t \rightarrow 1$).
% , leading to oversmoothed outputs.  
Although we find that this adjustment helps in long-step samplers, improving the quality of the generated structures, it becomes harmful when using only a few steps: larger $\eta$ values lead to unstable updates and degraded accuracy. Our experiments show that $\eta = 1.0$ is essential for few-step sampling. Figure~\ref{fig:eta-study-bar} illustrates this interaction.

\begin{figure}[h!]
  \centering
  \includegraphics[width=.95\textwidth]{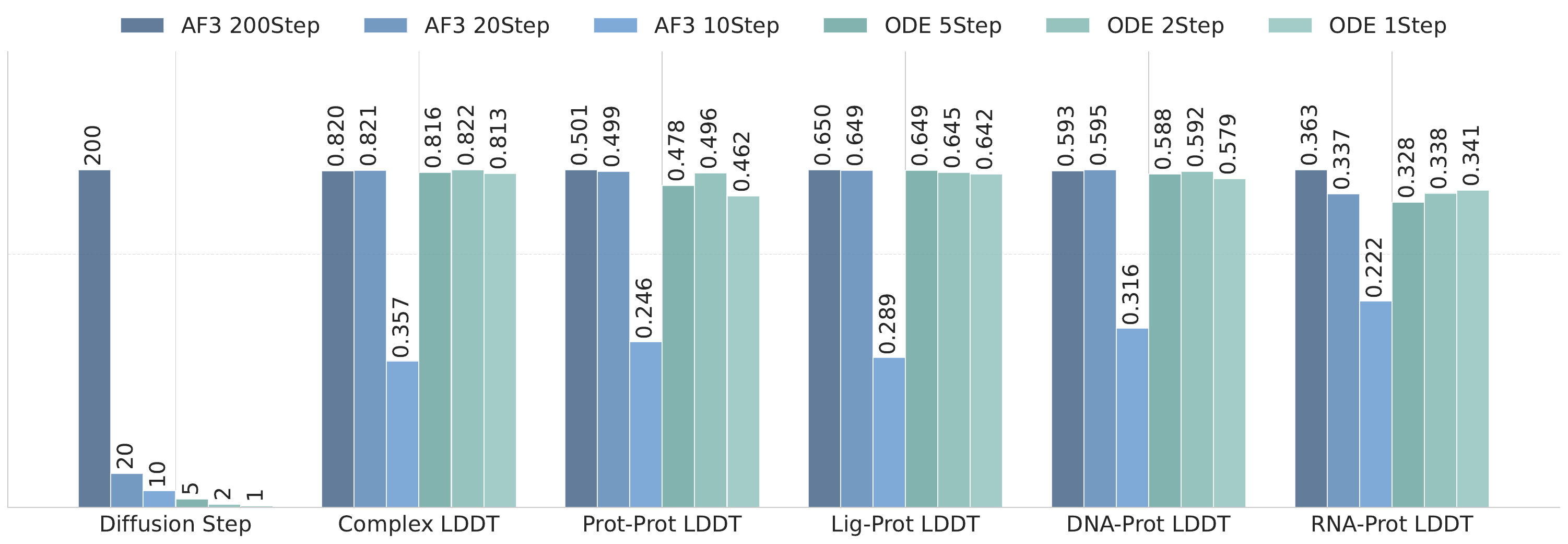}
  \caption{
    \textbf{Model performance across different samplers.}
    AF3 sampler uses $\eta = 1.5$, $\lambda = 1.003$, $\gamma_0 = 0.8$; ODE sampler uses $\eta = 1.0$, $\lambda = 1.0$, $\gamma_0 = 0$. 
  }
  \label{fig:sampler}
\end{figure}

Figure~\ref{fig:sampler} compares the default AF3 sampler ($\eta=1.5, \lambda=1.003, \gamma_0=0.8$) and the modified ODE sampler ($\eta=1.0, \lambda=1.0, \gamma_0=0$) across different step counts. The default configuration performs well with 200 or 20 steps, but collapses when reduced below 10, often failing entirely at 5 steps. In contrast, the ODE configuration remains stable even with only 1 or 2 steps. For example, the 2-step ODE sampler achieves 0.645 LDDT on ligand–protein interfaces, nearly matching the 0.65 score of the 200-step baseline, while the 1-step variant still achieves a reasonable 0.64.

While few-step inference achieves strong average performance, we observe occasional structural issues, such as reduced physical plausibility or atomic clashes. We provide illustrative examples in Section~\ref{sec:case-study}.

For completeness, we also evaluate models trained with flow matching in Section~\ref{sec:experiment}, showing similarly robust behavior.
The studies in this section was conducted using a small-scale Protenix model, but we believe the observed trends generalize well across model sizes and to other AF3-style models.

\subsection{Compress Redundant Blocks in the Protenix Model}

\begin{table}[h!]
    \centering
    \scalebox{0.85}{
    \begin{tabular}{cc|c|ccccc}
    \toprule
    \#pairformer & \#transformer & Mode & Prot-Prot & Lig-Prot & DNA-Prot & RNA-Prot & Intra-Prot \\
    \hline 
    48 & 24 & - & 0.501 & 0.650 & 0.593 & 0.363 & 0.844\\
    \hline
    44 & 24 & Zero-shot & 0.484 & 0.631 & 0.575 & 0.344 & 0.828\\
    44 & 24 & finetuned & 0.500 & 0.652 & 0.592 & 0.360 & 0.843 \\
    \bottomrule
    \end{tabular}}
    \caption{\textbf{The smaller model can achieve similar performance with further finetuning.} We fine-tune the model with batch size 256, learning rate $10^{-3}$ and 10K iterations. 
    % \cy{TODO: add more config results, re-new results on the V8 checkpoint}
    }
    \label{tab:blocks}
\end{table}

\begin{wrapfigure}{r}{0.4\textwidth}
    \centering
    \vspace{-20pt}
    \includegraphics[width=1\linewidth]{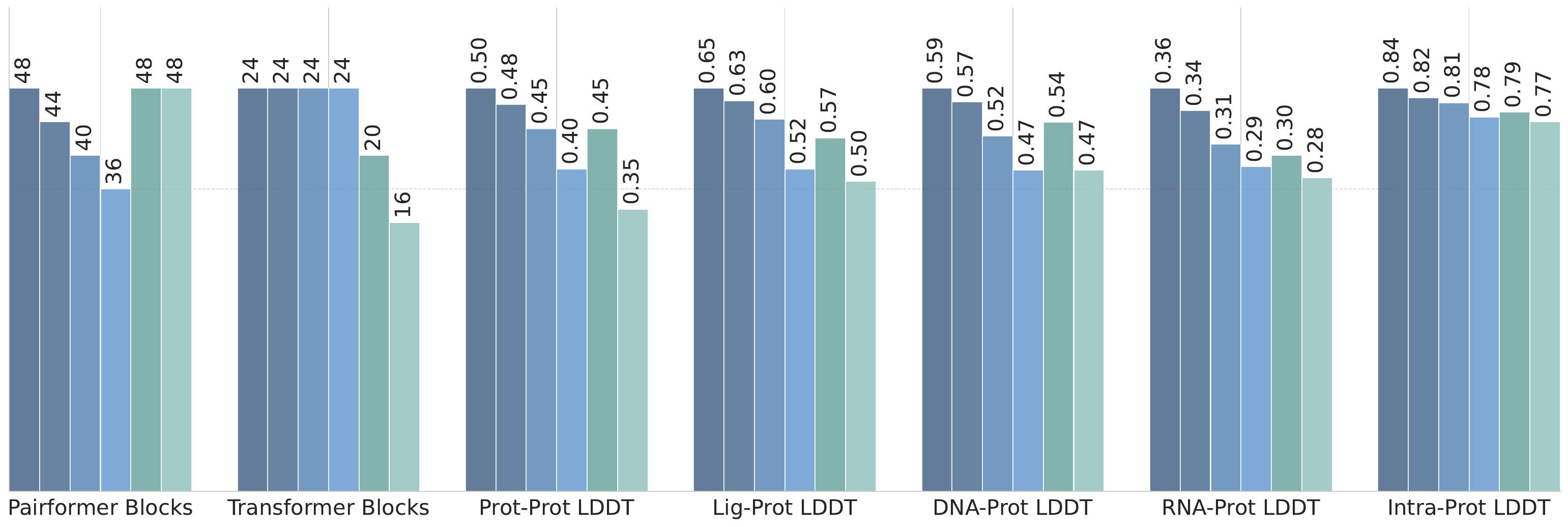}
    \vspace{-20pt}
    \caption{
    \textbf{Post-training pruning performance on Protenix.} Removing certain blocks without finetuning leads to only a slight drop in performance compared to the full-size model.}
    % \vspace{-30pt}
    \label{fig:block}
\end{wrapfigure}

The architecture of AF3 comprises 48 Pairformer blocks for modeling pairwise token distances and 24 Diffusion Transformer blocks for generating 3D coordinates. 
While these deep networks excel at capturing complex structural dependencies, their computational cost introduces substantial overhead and raises questions about architectural redundancy—specifically, whether all layers contribute meaningfully to the final prediction. Similar observations have been reported in prior work \citep[e.g.][]{zhang2022all, cheng2024survey, lee2019wide, ye2020good}. 
% Our analysis reveals that a significant subset of these blocks has minimal impact on prediction performance.
Specifically, we performed block-wise ablations by progressively removing early-stage blocks in the Pairformer network. Removing the first 4 Pairformer blocks resulted in only a slight loss in structural accuracy on benchmark datasets, as shown in Figure~\ref{fig:block}. In Table~\ref{tab:blocks}, we further observe that, with finetuning of the remaining model parameters, removing several blocks does not degrade performance. 
This finding motivates us to adopt a more compact architectural configuration.

Based on this findings, we tried two different training schemes, 1) pruning these blocks and finetune the model, 
2) directly setting a smaller architecture configuration and train from scratch. 
First, through a search over model architecture configurations, we found that the Protenix-Mini architecture yields a favorable trade-off between LDDT scores and efficiency, as listed in Figure \ref{fig:flops} and Table \ref{tab:appendix:architecture-config}.
We then performed block-wise pruning on the Pairformer blocks. Notably, dropping the first 8 Pairformer blocks resulted in around 10\% drop in complex LDDT (e.g., from 0.80 to 0.72) after 800 iterations of fine-tuning. We dropped the 8 Pairformer blocks closest to the inputs, fine-tuned the remaining model parameters, and obtained the Protenix-Tiny model. Detailed training configurations are listed in Table \ref{tab:appendix:architecture-config}.

\subsection{pLM Based Protenix-Mini}
Given the substantial computational burdens imposed by both MSA search algorithms and MSA computation modules--often requiring extensive memory and parallel processing for MSA tasks---we propose a strategic substitution:
integrating a pre-trained protein language model to supplant the traditional MSA pipeline.

For the model architecture, 
we first convert the input sequence into embeddings, then use a linear transformation layer to convert the embedding vectors dimension. 
In our implementation, we leverage the ESM2-3B model \citep{lin2023evolutionary} to generate contextualized sequence embeddings, which are then added to the $\textrm{s\_inputs}$ in the the input embedder module of the structure prediction pipeline. We nullify the MSA contribution by passing empty MSA input dictionary during both training and inference.
 
For model training,
to facilitate knowledge transfer from MSA-based models to the ESM-driven architecture, we introduce a hybrid training strategy.
During each training iteration, the model randomly selects either the MSA module or the ESM features with 50\% probability, and 
all the core components (e.g., the atom encoder, the atom decoder, transformer blocks, diffusion transformer blocks) are shared between the two pathways, ensuring consistency in feature representation.
This framework can be viewed as implicitly performing knowledge distillation, as the ESM module learns to mimic the MSA module's output distributions without explicit supervision, leading to compact yet informative embeddings.
During inference,  the MSA module is entirely omitted and we only use the EDM module. % inference.
% This can be viewed as some soft knowledge distillation technique.

\section{Experimental Results}
\label{sec:experiment}
% \cy{in all the figures, add error bars}

\subsection{Model Configuration and Training Details}
% Describe the model config:
% \cy{in all the figures, add description about LDDT}
Following the above analysis, we train a Protenix-Mini model with $16$ pairformer blocks, $8$ diffusion transformer blocks, $1$ MSA module block, $1$ atom decoder block and $1$ atom encoder block. We keep all the other training and model architecture configurations the same as the open-sourced Protenix model \citep{bytedance2025protenix} and project \footnote{\url{https://github.com/bytedance/Protenix}}.
We trained the model with both the EDM and flow matching framework, and notice that these two approaches come to similar performance. 
For Protenix-Mini, we applied a batch size of 64, 200K iterations ans a learning rate of $10^{-3}$.
% \cy{add more training details for each model, e.g., batch size, number of iterations}
% 
Once we had a well-trained Protenix-Mini model,
we further pruned the first 8 Pairformer blocks closest to the input and fine-tune the resmaining weights in the network. 
We therefore get a smaller network with $8$ Pairformer blocks and marks the model as the Protenix-Tiny model.
After pruning, we further tune the model with batch size 64 and 100K iterations.
We also loaded the Mini model weights, set MSA as empty, added the ESM2 representation as part of the inputs, and got the Protenix-Mini-ESM model for an additional 100K-iteration training. 
The other training details have been discussed in the sections above.
During inference, we use $4$ cycles for the backbone model and $2$-step ODE sampler for the diffusion model as mentioned in Alg.~\ref{alg:rectflow-sample} to save computation.

In all subsequent experiments, we report complex and interface LDDT as the evaluation metrics.
We evaluate the model performance on a subset of the recentPDB, with fewer than 768 tokens.

\begin{table}[t]
    \centering
    \begin{tabular}{c|cccccc}
    \toprule
    Model / Domain  & Prot-Prot & Lig-Prot & DNA-Prot & RNA-Prot & Intra-Prot & Complex LDDT \\
    \hline 
    Protenix & 0.501 & 0.650 & 0.593 & 0.363 & 0.844 & 0.820 \\
    \hline
    % Protenix-Mini & 0.510 & 0.650 & 0.579 & 0.344 & 0.833 & 0.816\\
    Protenix-Mini & 0.490 & 0.622 & 0.584 & 0.339 & 0.818 & 0.802\\
    Protenix-Tiny & 0.428 & 0.594 & 0.567 & 0.338 & 0.798 & 0.783\\
    \bottomrule
    \end{tabular}
    \caption{\textbf{Model Performance on RecentPDB$\leq 768$ test set.} We average the scores of random samples without leveraging confidence scores for sample selection.  We report interface LDDT for different interface types. 
    % xw\cy{to be updated, we get better tiny results.}
    % The moitivation is to  compare the backbone structure generation performance here.
    }
    \label{tab:main_results}
\end{table}
\begin{table}[t]
    \centering
    \begin{tabular}{c|cccccc}
    \toprule
    Model / Domain  & Prot-Prot & Lig-Prot & DNA-Prot & RNA-Prot & Intra-Prot & Complex LDDT \\
    \hline 
    Protenix & 0.501 & 0.650 & 0.593 & 0.363 & 0.844 & 0.820 \\
    \hline
    % Protenix-Mini & 0.510 & 0.650 & 0.579 & 0.344 & 0.833 & 0.816\\
    Protenix-Mini & 0.490 & 0.622 & 0.584 & 0.339 & 0.818 & 0.802\\
    Protenix-Mini-ESM & 0.408 & 0.568 & 0.602 & 0.294 & 0.792 & 0.775\\
    \bottomrule
    \end{tabular}
    \caption{\textbf{ESM Model Performance on RecentPDB($\leq 768$) test set.} We average scores across multiple samples  without leveraging confidence scores for sample selection. 
    % We report interface LDDT for different interface types. 
    % The moitivation is to  compare the backbone structure generation performance here.
    }
    \label{tab:esm_results}
\end{table}

\begin{table}[tb]
    \centering
    \begin{tabular}{c|cccccc}
    \toprule
    Model / Domain  & Prot-Prot & Lig-Prot & DNA-Prot & RNA-Prot & Intra-Prot & Complex LDDT \\
    \hline 
    Protenix & 0.501 & 0.650 & 0.593 & 0.363 & 0.844 & 0.820 \\
    \hline
    % Protenix-Mini & 0.510 & 0.650 & 0.579 & 0.344 & 0.833 & 0.816\\
    PTX-Mini  & 0.490 & 0.622 & 0.584 & 0.339 & 0.818 & 0.802\\
    PTX-Mini-Flow & 0.488 & 0.619 & 0.596 & 0.348 & 0.814 & 0.797 \\
    \hline
    % Protenix-Mini & 0.510 & 0.650 & 0.579 & 0.344 & 0.833 & 0.816\\
    PTX-Mini-ESM & 0.408 & 0.568 & 0.602 & 0.294 & 0.792 & 0.775\\
    PTX-Mini-ESM-Flow & 0.405 & 0.577 & 0.585 & 0.297 & 0.787 & 0.772\\
    \hline
    Protenix-Tiny & 0.428 & 0.594 & 0.567 & 0.338 & 0.798 & 0.783\\
    PTX-Tiny-Flow & 0.422 & 0.594 & 0.569 & 0.338 & 0.806 & 0.780\\
    \bottomrule
    \end{tabular}
    \caption{\textbf{Flow Model Performance on RecentPDB($\leq 768$) test set.} 
    % We average scores of random samples without leveraging confidence scores for sample selection. 
    `Flow' denotes the model is trained with flow matching loss.
    % We report interface LDDT for different interface types. 
    % The moitivation is to  compare the backbone structure generation performance here.
    }
    \label{tab:flow_results}
\end{table}

\begin{table}[t]
    \centering
    % \vspace{-20pt}
    \begin{tabular}{c|cc}
    \toprule
    Model & Ligand RMSD & Success Rate (\%)\\
    \hline 
    Protenix & 1.95 & 80.0\\
    \hline
    Protenix-Mini & 2.22 & 72.7\\
    Protenix-Tiny & 2.28 & 69.8\\
    \bottomrule
    \end{tabular}
    \caption{\textbf{Model Performance on PB($\leq 768$) test set.} We random sample and take the average scores without leveraging confidence scores for sample selection. Here, the success rate is defined as RMSD $\leq 2$.
    % The moitivation is to  compare the backbone structure generation performance here.
    }
    % \vspace{-20pt}
    \label{tab:pb}
\end{table}

\subsection{Results}
The model performances on two dataset, RecentPDB($\leq 768$) and Posebusters($\leq 768$), 
are reported. 
As demonstrated in Table \ref{tab:main_results}, 
% we compare different model performances on the RecentPDB test dataset, with LDDT and interface LDDT as evaluation metrics.
the Mini model exhibits slightly lower performance than the standard Protenix model on both complex-level and interface-level LDDT metrics.
The Tiny model, on the other hand, get around 2\% to 3\% performance drop on different interface types, while reducing the inference computational cost by around 85\% compared to the Protenix model as mentioned in Figure \ref{fig:flops}.

On Posebusters, the average ligand RMSD and average success rate (ligand RMSD $\leq$ 2) are reported as evaluation metrics.
% We compare the model performance on the Posebusters, and 
% we report the average ligand RMSD and average success rate (ligand RMSD $\leq$ 2) on this dataset. 
Table \ref{tab:pb} shows that the Mini model performs nearly as well as the full Protenix model, while the Tiny model yields marginally worse results.

\subsubsection{ESM Model Performance}
Protenix-Mini-ESM model performance is demonstrated in Table \ref{tab:esm_results}.
The ESM-based model get slightly worse performance on most of the test metrics.
For example, the complex LDDT drops from $0.8$ to $0.775$. 
However, the protein protein interface performance has a large-margin drop. 
While Protenix-Mini gets around $0.5$ interface LDDT, ESM version model gets around $0.4$ interface LDDT score.
This over 10\% performance degradation highlights the critical role of paired Multiple Sequence Alignments (MSAs) in protein-protein interface prediction. Going forward, we intend to develop structure-aware protein language models to further enhance predictive performance and accuracy.

\subsubsection{Flow Matching Model Performance}
Beyond the EDM framework, we also emplyed a flow matching framework.
The primary distinction is replacing the $x$ (output) prediction network with a velocity prediction network, alongside differences in the $t$-weighting function used during training.

% Based on the fact that, ODE sampler yields more efficient performance, one natural and regular choice is applying flow matching / rectified flow loss to training the model parameters.
For flow matching training, we also include the same losses as the EDM framework.  First, we use 
$
\mathcal{L}_\mathrm{mse} = \| x_1 - x_0 - \mathrm{Diffuser}(x_t, t \mid C) \|^2,
$
to learn the velocity. 
We compute additional losses (e.g., smooth LDDT loss, bond loss) using the estimation %$\hat{x}_1$,
$
\hat{x}_1 = x_t + (1 - t) \cdot \mathrm{Diffuser}(x_t, t \mid C).
$
In practice, we randomly sample $t$ from the $\mathrm{Beta}(2.5, 2.5)$ distribution during training.
% Mathematically, the Beta distribution with parameters $\alpha = \beta = 2.5$ has a probability density function (PDF) 
The density peaks around $t = 0.5$, and assigns lower weight near $t=0$ and $t=1$, 
serving two purposes: (1) prioritizing learning noise correction in intermediate diffusion stages (where data balances noise and structure, critical for fine-grained details); (2) mitigating overfitting to trivial cases (e.g., near-identity mappings) and unstable predictions in highly noisy regimes.
% This sampling strategy serves two key purposes, 
% By up-weighting $t=0.5$ region, the model prioritizes learning noise correction in the intermediate diffusion stages, where the data exhibits a balance between noise and structure—critical for capturing fine-grained structural details.
% Down-weighting $t=0$ and $t=1$ mitigates overfitting to trivial cases (e.g., near-identity mappings) and unstable predictions in highly noisy regimes. 

We compare the flow matching and EDM model performance in several different settings. 
As listed in Table \ref{tab:flow_results}, flow matching and EDM framework achieves similar performance under different settings. 
For example, the score differences between PTX-Mini and PTX-Mini-Flow across domains range from -0.005 (RNA-Prot) to +0.009 (DNA-Prot), with Complex LDDT differing by only 0.013.
These negligible variations indicate that the flow matching and EDM do not play different impacts on the model's performance across interaction types.
% Considering that the key difference between these models is how to weight and sample different time $t$, 
% suggesting that the core architecture dominates predictive power compared to different diffusion structure generation framework. 

\begin{table}[t]
    \centering
    \begin{tabular}{l|l|ccc|ccc}
    \toprule
    & & \multicolumn{3}{c|}{Intra-Protein Clash } & \multicolumn{3}{c}{Protein-Ligand Clash} \\
    \cline{3-8}
    & & $\exists$ & $\forall 1\times5$ & $\forall 5\times5$ & $\exists$ & $\forall 1\times5$ & $\forall 5\times5$ \\
    \hline
    Protenix & 20-step AF3 & 0 & 0 & 0 & 6 & 0 & 0 \\
    Protenix & 10-step ODE & 0 & 0 & 0 & 10 & 0 & 0 \\
    Protenix & 2-step ODE  & 11 & 0 & 0 & 23 & 0 & 0  \\
    \hline
    Mini & 20-step AF3 & 0 & 0 & 0 & 6 & 0 & 0 \\
    Mini & 10-step ODE & 0 & 0 & 0 & 11 & 0 & 0 \\
    Mini & 2-step ODE & 7 & 0 & 0 & 22 & 0 & 0 \\
    \hline
     Mini-ESM & 200-step AF3  & 0 & 0 & 0  & 7 & 0 & 0 \\
    Mini-ESM & 20-step AF3 & 0 & 0 & 0 & 7 & 0 & 0 \\
    Mini-ESM & 2-step ODE & 17 & 0 & 0 & 22 & 1 & 0 \\
    Mini-EDM-Flow & 2-step ODE & 7 & 0 & 0  & 21 & 1 & 0 \\
    \bottomrule
    \end{tabular}
    \caption{\textbf{Number of Clashes on PB}. We generate $5\times5$ samples and valid whether they have clashes. $\exists$ means there is clashed case in $5\times5$ samples, while $\forall 5\times5$ means all $5\times5$ samples have clash. 
    % \cy{re-do: AF3 clash only consider polymer chains. conside ligand-polymer distance, we get mini-sde 6 clashed cases, mini-ode 22 clashed cases. }
    % The moitivation is to  compare the backbone structure generation performance here.
    }
    \label{tab:clash}
\end{table}

\begin{table}[t]
    \centering
    \begin{tabular}{c|cccccc}
    \toprule
    Model / Domain  & Prot-Prot & Lig-Prot & DNA-Prot & RNA-Prot & Intra-Prot & Complex LDDT \\
    \hline 
    20-Step AF3 & 0.205 & 0.179 & 0.083 & 0.214 & 0.046 & 0.038 \\
    \hline
    2-Step ODE & 0.209 & 0.181 & 0.096 & 0.257 & 0.050 & 0.044 \\
    \bottomrule
    \end{tabular}
    \caption{\textbf{Protenix Model Sampler Diversity on RecentPDB($\leq 768$) test set.} We illustrate the marked disparity between the top-performing and worst cases across $5 \times 5$ samples on the Protenix model. Notably, our analyses confirm that the ODE sampler preserves structural diversity without degradation. 
    % \cy{add more models}
    % The moitivation is to  compare the backbone structure generation performance here.
    }
    \label{tab:diversity}
\end{table}

% \cy{title shouldn't be the last line}
We quantitatively compared the difference between the 2-step ODE and multi-step AF3 sampler.
% , to understand their difference in the predicted structures. In the following studies, 
For these studies, we sample 5 different seeds, and for each seed, we generate 5 diffusion samples.
First, we observed that, when applying to the Mini EDM model, both samplers suffer from the clash problem.
As demonstrated in Table \ref{tab:clash}, the Mini model is more prone to generating structures with clashes, though generating additional samples mitigates this in practice. For the full Protenix model, the 2-step ODE sampler still exhibits clashes, whereas the multi-step AF3 sampler performs better in this regard.

\subsubsection{Case Study and Analyses}
\label{sec:case-study}
\begin{wrapfigure}{r}{0.22\textwidth}
    \centering
    \vspace{-40pt}
    \includegraphics[width=1\linewidth]{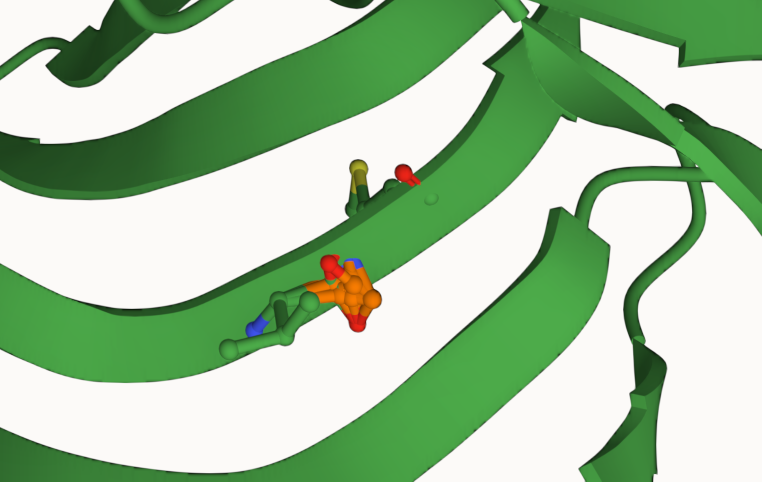}
    \vspace{-10pt}
    \caption{
    \textbf{2-step ODE gets collapsed results sometimes.} E.g., `7bnh'.}
    \vspace{-12.5pt}
    \label{fig:bas-case}
\end{wrapfigure}

% For the large model, we observe that different samplers do not have this problem.
We also visually analyzed differences between multi-step and few-step samplers.
Compared to the multi-step AF3 sampler, 2-step ODE sampler generated more collapsed ligands, as shown in Figure \ref{fig:bas-case}. 
Notably, increasing the ODE sampling steps to 10 effectively mitigates this issue, suggesting that the problem may cause by the underfitting in the region when $t \leq 0.5$ for the diffusion model. In practice, we notice that set the first 5 ODE step with $\eta=1$ and the last 5 ODE step with $\eta=1.5$ comes to high-quality results without clashes.
% It suggests that,  Once $t \leq 0.5$ region has better fitness, we can solve this problem. 
% \cy{do more-step ODE inference, to check whether we have the same problem}

We also assessed diversity across samples and found that the disparity between best and worst samples is nearly identical for the AF3 and ODE samplers. 
Case studies, as quantified in Table \ref{tab:diversity}, further confirm comparable diversity profiles.
These findings underscore a key finding: while few-step ODE sampling may yield occasional structural anomalies for some cases, it preserves sampling diversity and can achieve competitive performance with appropriate step-size optimization.
% Based on all these findings, we come to the conclusion that, the few-step sampler may cause some bad cases, but the few-step ODE samplers do not lose diversity and can come to good performance. 

\section{Conclusions and Future Directions}
We present Protenix-Mini, a lightweight framework merging compact architecture design with a 2-step ODE sampler to achieve efficient biomolecular structure prediction, demonstrating minimal performance loss (1\% to 5\%) on key benchmarks. By pruning redundant transformer components and leveraging stable loss functions, we showcase the feasibility of balancing accuracy with inference speed, addressing a critical gap in deploying diffusion models for real-world applications.
We also propose a ESM-based variant to replace the MSA search module for better effectiveness and efficiency trade-off.

Moving forward, we pursue several complementary strategies to further enhance efficiency, including 1) exploring sparse or adaptive attention architectures to optimize transformer backbones for large biomolecular complexes, mitigating their quadratic computational complexity, 2) searching for better trade-off in detailed architecture configurations, 3) pretraining pLM with structure information. These efforts aim to unlock scalable, low-latency structure prediction for long sequences and multi-component systems, driving advancements in drug discovery and synthetic biology.

\nocite{langley00}

\bibliography{example_paper}

\begin{thebibliography}{27}
\providecommand{\natexlab}[1]{#1}
\providecommand{\url}[1]{\texttt{#1}}
\expandafter\ifx\csname urlstyle\endcsname\relax
  \providecommand{\doi}[1]{doi: #1}\else
  \providecommand{\doi}{doi: \begingroup \urlstyle{rm}\Url}\fi

\bibitem[Abramson et~al.(2024)Abramson, Adler, Dunger, Evans, Green, Pritzel, Ronneberger, Willmore, Ballard, Bambrick, et~al.]{abramson2024accurate}
Abramson, J., Adler, J., Dunger, J., Evans, R., Green, T., Pritzel, A., Ronneberger, O., Willmore, L., Ballard, A.~J., Bambrick, J., et~al.
\newblock Accurate structure prediction of biomolecular interactions with alphafold 3.
\newblock \emph{Nature}, 630\penalty0 (8016):\penalty0 493--500, 2024.

\bibitem[Cao et~al.(2022)Cao, Coventry, Goreshnik, Huang, Sheffler, Park, Jude, Markovi{\'c}, Kadam, Verschueren, et~al.]{cao2022design}
Cao, L., Coventry, B., Goreshnik, I., Huang, B., Sheffler, W., Park, J.~S., Jude, K.~M., Markovi{\'c}, I., Kadam, R.~U., Verschueren, K.~H., et~al.
\newblock Design of protein-binding proteins from the target structure alone.
\newblock \emph{Nature}, 605\penalty0 (7910):\penalty0 551--560, 2022.

\bibitem[Cheng et~al.(2024)Cheng, Zhang, and Shi]{cheng2024survey}
Cheng, H., Zhang, M., and Shi, J.~Q.
\newblock A survey on deep neural network pruning: Taxonomy, comparison, analysis, and recommendations.
\newblock \emph{IEEE Transactions on Pattern Analysis and Machine Intelligence}, 2024.

\bibitem[Cheng et~al.(2023)Cheng, Novati, Pan, Bycroft, {\v{Z}}emgulyt{\.e}, Applebaum, Pritzel, Wong, Zielinski, Sargeant, et~al.]{cheng2023accurate}
Cheng, J., Novati, G., Pan, J., Bycroft, C., {\v{Z}}emgulyt{\.e}, A., Applebaum, T., Pritzel, A., Wong, L.~H., Zielinski, M., Sargeant, T., et~al.
\newblock Accurate proteome-wide missense variant effect prediction with alphamissense.
\newblock \emph{Science}, 381\penalty0 (6664):\penalty0 eadg7492, 2023.

\bibitem[Jumper et~al.(2021)Jumper, Evans, Pritzel, Green, Figurnov, Ronneberger, Tunyasuvunakool, Bates, {\v{Z}}{\'\i}dek, Potapenko, et~al.]{jumper2021highly}
Jumper, J., Evans, R., Pritzel, A., Green, T., Figurnov, M., Ronneberger, O., Tunyasuvunakool, K., Bates, R., {\v{Z}}{\'\i}dek, A., Potapenko, A., et~al.
\newblock Highly accurate protein structure prediction with alphafold.
\newblock \emph{Nature}, 596\penalty0 (7873):\penalty0 583--589, 2021.

\bibitem[Karras et~al.(2022)Karras, Aittala, Aila, and Laine]{karras2022elucidating}
Karras, T., Aittala, M., Aila, T., and Laine, S.
\newblock Elucidating the design space of diffusion-based generative models.
\newblock \emph{Advances in neural information processing systems}, 35:\penalty0 26565--26577, 2022.

\bibitem[Karras et~al.(2024)Karras, Aittala, Kynk{\"a}{\"a}nniemi, Lehtinen, Aila, and Laine]{karras2024guiding}
Karras, T., Aittala, M., Kynk{\"a}{\"a}nniemi, T., Lehtinen, J., Aila, T., and Laine, S.
\newblock Guiding a diffusion model with a bad version of itself.
\newblock \emph{Advances in Neural Information Processing Systems}, 37:\penalty0 52996--53021, 2024.

\bibitem[Langley(2000)]{langley00}
Langley, P.
\newblock Crafting papers on machine learning.
\newblock In Langley, P. (ed.), \emph{Proceedings of the 17th International Conference on Machine Learning (ICML 2000)}, pp.\  1207--1216, Stanford, CA, 2000. Morgan Kaufmann.

\bibitem[Lee et~al.(2019)Lee, Xiao, Schoenholz, Bahri, Novak, Sohl-Dickstein, and Pennington]{lee2019wide}
Lee, J., Xiao, L., Schoenholz, S., Bahri, Y., Novak, R., Sohl-Dickstein, J., and Pennington, J.
\newblock Wide neural networks of any depth evolve as linear models under gradient descent.
\newblock \emph{Advances in neural information processing systems}, 32, 2019.

\bibitem[Lin et~al.(2023)Lin, Akin, Rao, Hie, Zhu, Lu, Smetanin, Verkuil, Kabeli, Shmueli, et~al.]{lin2023evolutionary}
Lin, Z., Akin, H., Rao, R., Hie, B., Zhu, Z., Lu, W., Smetanin, N., Verkuil, R., Kabeli, O., Shmueli, Y., et~al.
\newblock Evolutionary-scale prediction of atomic-level protein structure with a language model.
\newblock \emph{Science}, 379\penalty0 (6637):\penalty0 1123--1130, 2023.

\bibitem[Lipman et~al.(2022)Lipman, Chen, Ben-Hamu, Nickel, and Le]{lipman2022flow}
Lipman, Y., Chen, R.~T., Ben-Hamu, H., Nickel, M., and Le, M.
\newblock Flow matching for generative modeling.
\newblock \emph{arXiv preprint arXiv:2210.02747}, 2022.

\bibitem[Liu(2022)]{liu2022rectified}
Liu, Q.
\newblock Rectified flow: A marginal preserving approach to optimal transport.
\newblock \emph{arXiv preprint arXiv:2209.14577}, 2022.

\bibitem[Liu et~al.(2023{\natexlab{a}})Liu, Gong, et~al.]{liuflow}
Liu, X., Gong, C., et~al.
\newblock Flow straight and fast: Learning to generate and transfer data with rectified flow.
\newblock In \emph{The Eleventh International Conference on Learning Representations}, 2023{\natexlab{a}}.

\bibitem[Liu et~al.(2023{\natexlab{b}})Liu, Wu, Zhang, Gong, Ping, and Liu]{liu2023flowgrad}
Liu, X., Wu, L., Zhang, S., Gong, C., Ping, W., and Liu, Q.
\newblock Flowgrad: Controlling the output of generative odes with gradients.
\newblock In \emph{Proceedings of the IEEE/CVF Conference on Computer Vision and Pattern Recognition}, pp.\  24335--24344, 2023{\natexlab{b}}.

\bibitem[Liu et~al.(2023{\natexlab{c}})Liu, Zhang, Ma, Peng, et~al.]{liu2023instaflow}
Liu, X., Zhang, X., Ma, J., Peng, J., et~al.
\newblock Instaflow: One step is enough for high-quality diffusion-based text-to-image generation.
\newblock In \emph{The Twelfth International Conference on Learning Representations}, 2023{\natexlab{c}}.

\bibitem[Lu et~al.(2022)Lu, Zhou, Bao, Chen, Li, and Zhu]{lu2022dpm}
Lu, C., Zhou, Y., Bao, F., Chen, J., Li, C., and Zhu, J.
\newblock Dpm-solver: A fast ode solver for diffusion probabilistic model sampling in around 10 steps.
\newblock \emph{Advances in Neural Information Processing Systems}, 35:\penalty0 5775--5787, 2022.

\bibitem[Ma et~al.(2025)Ma, Tong, Jia, Hu, Su, Zhang, Yang, Li, Jaakkola, Jia, et~al.]{ma2025inference}
Ma, N., Tong, S., Jia, H., Hu, H., Su, Y.-C., Zhang, M., Yang, X., Li, Y., Jaakkola, T., Jia, X., et~al.
\newblock Inference-time scaling for diffusion models beyond scaling denoising steps.
\newblock \emph{arXiv preprint arXiv:2501.09732}, 2025.

\bibitem[Pacesa et~al.(2024)Pacesa, Nickel, Schellhaas, Schmidt, Pyatova, Kissling, Barendse, Choudhury, Kapoor, Alcaraz-Serna, et~al.]{pacesa2024bindcraft}
Pacesa, M., Nickel, L., Schellhaas, C., Schmidt, J., Pyatova, E., Kissling, L., Barendse, P., Choudhury, J., Kapoor, S., Alcaraz-Serna, A., et~al.
\newblock Bindcraft: one-shot design of functional protein binders.
\newblock \emph{bioRxiv}, pp.\  2024--09, 2024.

\bibitem[Song et~al.(2020)Song, Meng, and Ermon]{song2020denoising}
Song, J., Meng, C., and Ermon, S.
\newblock Denoising diffusion implicit models.
\newblock \emph{arXiv preprint arXiv:2010.02502}, 2020.

\bibitem[Team et~al.(2025)Team, Chen, Zhang, Lu, Ma, Guan, Gong, Yang, Zhang, Zhang, Wu, Zhou, Yang, Liu, Wang, Shi, Shi, and Xiao]{bytedance2025protenix}
Team, B. A.~A., Chen, X., Zhang, Y., Lu, C., Ma, W., Guan, J., Gong, C., Yang, J., Zhang, H., Zhang, K., Wu, S., Zhou, K., Yang, Y., Liu, Z., Wang, L., Shi, B., Shi, S., and Xiao, W.
\newblock Protenix - advancing structure prediction through a comprehensive alphafold3 reproduction.
\newblock \emph{bioRxiv}, pp.\  2025--01, 2025.
\newblock \doi{10.1101/2025.01.08.631967}.
\newblock URL \url{https://www.biorxiv.org/content/early/2025/01/11/2025.01.08.631967}.

\bibitem[team et~al.(2024)team, Boitreaud, Dent, McPartlon, Meier, Reis, Rogozhonikov, and Wu]{chai2024chai}
team, C.~D., Boitreaud, J., Dent, J., McPartlon, M., Meier, J., Reis, V., Rogozhonikov, A., and Wu, K.
\newblock Chai-1: Decoding the molecular interactions of life.
\newblock \emph{BioRxiv}, pp.\  2024--10, 2024.

\bibitem[Wang et~al.(2023)Wang, Fu, Du, Gao, Huang, Liu, Chandak, Liu, Van~Katwyk, Deac, et~al.]{wang2023scientific}
Wang, H., Fu, T., Du, Y., Gao, W., Huang, K., Liu, Z., Chandak, P., Liu, S., Van~Katwyk, P., Deac, A., et~al.
\newblock Scientific discovery in the age of artificial intelligence.
\newblock \emph{Nature}, 620\penalty0 (7972):\penalty0 47--60, 2023.

\bibitem[Watson et~al.(2023)Watson, Juergens, Bennett, Trippe, Yim, Eisenach, Ahern, Borst, Ragotte, Milles, et~al.]{watson2023novo}
Watson, J.~L., Juergens, D., Bennett, N.~R., Trippe, B.~L., Yim, J., Eisenach, H.~E., Ahern, W., Borst, A.~J., Ragotte, R.~J., Milles, L.~F., et~al.
\newblock De novo design of protein structure and function with rfdiffusion.
\newblock \emph{Nature}, 620\penalty0 (7976):\penalty0 1089--1100, 2023.

\bibitem[Wohlwend et~al.(2024)Wohlwend, Corso, Passaro, Reveiz, Leidal, Swiderski, Portnoi, Chinn, Silterra, Jaakkola, et~al.]{wohlwend2024boltz}
Wohlwend, J., Corso, G., Passaro, S., Reveiz, M., Leidal, K., Swiderski, W., Portnoi, T., Chinn, I., Silterra, J., Jaakkola, T., et~al.
\newblock Boltz-1: Democratizing biomolecular interaction modeling.
\newblock \emph{bioRxiv}, pp.\  2024--11, 2024.

\bibitem[Yang et~al.(2023)Yang, Zhang, Song, Hong, Xu, Zhao, Zhang, Cui, and Yang]{yang2023diffusion}
Yang, L., Zhang, Z., Song, Y., Hong, S., Xu, R., Zhao, Y., Zhang, W., Cui, B., and Yang, M.-H.
\newblock Diffusion models: A comprehensive survey of methods and applications.
\newblock \emph{ACM Computing Surveys}, 56\penalty0 (4):\penalty0 1--39, 2023.

\bibitem[Ye et~al.(2020)Ye, Gong, Nie, Zhou, Klivans, and Liu]{ye2020good}
Ye, M., Gong, C., Nie, L., Zhou, D., Klivans, A., and Liu, Q.
\newblock Good subnetworks provably exist: Pruning via greedy forward selection.
\newblock In \emph{International Conference on Machine Learning}, pp.\  10820--10830. PMLR, 2020.

\bibitem[Zhang et~al.(2022)Zhang, Bengio, and Singer]{zhang2022all}
Zhang, C., Bengio, S., and Singer, Y.
\newblock Are all layers created equal?
\newblock \emph{Journal of Machine Learning Research}, 23\penalty0 (67):\penalty0 1--28, 2022.

\end{thebibliography}
\bibliographystyle{icml2025}

%%%%%%%%%%%%%%%%%%%%%%%%%%%%%%%%%%%%%%%%%%%%%%%%%%%%%%%%%%%%%%%%%%%%%%%%%%%%%%%
%%%%%%%%%%%%%%%%%%%%%%%%%%%%%%%%%%%%%%%%%%%%%%%%%%%%%%%%%%%%%%%%%%%%%%%%%%%%%%%
% APPENDIX
%%%%%%%%%%%%%%%%%%%%%%%%%%%%%%%%%%%%%%%%%%%%%%%%%%%%%%%%%%%%%%%%%%%%%%%%%%%%%%%
%%%%%%%%%%%%%%%%%%%%%%%%%%%%%%%%%%%%%%%%%%%%%%%%%%%%%%%%%%%%%%%%%%%%%%%%%%%%%%%
\newpage
\appendix
\onecolumn

\section{Training Details}
\label{app:training}

\subsection*{EDM Training Objectives}

Denote $x_0$ as random noise, and $x_1$ as the target structure, the EDM training objectives contain several terms in Alphafold3 (AF3) \citep{abramson2024accurate}, e.g., MSE loss, smooth LDDT loss and bond loss. 
For simplicity, we denote all the conditions for the diffusion model as $C$.
Mark $\hat{x}_1 = \mathrm{Diffuser}(\cdot | x_t, t; C)$, we have 
$$
\mathcal{L}_\mathrm{mse} = \| x_1 - \hat{x}_1\|^2 ~~\mathrm{and}~~
\mathcal{L}_\mathrm{bond} = \mathrm{mean}_{(l,m)} \bigg(   \| \overset{\rightarrow}{\hat{x}_1^l} - \overset{\rightarrow}{\hat{x}_1^m} \| - \| \overset{\rightarrow}{x_1^l} - \overset{\rightarrow}{x_1^m} \|\bigg)^2,
$$ 
where $l$ and $m$ denote the start and the end atom index for the bonds between the bonded ligand and parent chain, respectively.
For the smooth LDDT loss, an estimation for the LDDT metric, we take $\Delta x_{lm} \leftarrow \| \overset{\rightarrow}{\hat{x}_1^l} - \overset{\rightarrow}{\hat{x}_1^m} \|$ and $\Delta x_{lm}^{\mathrm{GT}}  \leftarrow \| \overset{\rightarrow}{x_1^l} - \overset{\rightarrow}{x_1^m} \|$ as input. 
After computing the distance difference for all pairs of atoms $\Delta_{lm} = \mathrm{abs}(\Delta x_{lm}^{\mathrm{GT}} - \Delta x_{lm})$, we smooth the values with
$$
\epsilon_{lm} \leftarrow \frac{1}{4} \bigg [ \sigma(1/2 - \Delta x_{lm} ) + \sigma(1 - \Delta x_{lm} ) +\sigma(2 - \Delta x_{lm} ) +\sigma(4 - \Delta x_{lm} )\bigg ],
$$
and compute the smooth LDDT loss as
$$
\mathcal{L}_\mathrm{SmoothLDDT} = 1 - \mathrm{mean}_{l \neq m} (w_{lm} \epsilon_{lm}) / \mathrm{mean}_{l \neq m} (w_{lm}),
$$
in which $w_{lm}$ denotes a weighting value determined by the entity type and radius.
We refer the readers to the AF3 for the details of the definition and computing for these losses (e.g., different atom weights, loss weights and scaling w.r.t. $t$, atom alignment and other details).

\iffalse
\subsubsection*{Flow Matching Training Objectives}
Except EDM framework, we also apply a flow matching framework.
Based on the fact that, ODE sampler yields more efficient performance, one natural and regular choice is applying flow matching / rectified flow loss to training the model parameters.
For flow matching training, we also include the same losses as the EDM framework. First, we have 
$
\mathcal{L}_\mathrm{mse} = \| x_1 - x_0 - g(\cdot | x_t, t; C) \|^2,
$
which learns the velocity. 
We further calculate the other losses, e.g., smooth LDDT loss, bond loss, with an estimated %$\hat{x}_1$,
$
\hat{x}_1 = x_t + (1 - t) g(\cdot | x_t, t; C).
$
In practice, we sample $t$ from the $\mathrm{Beta}(2.5, 2.5)$ distribution during training.
Mathematically, the Beta distribution with parameters $\alpha = \beta = 2.5$ has a probability density function (PDF) peak around $t = 0.5$, with reduced weight near $t=0$ and $t=1$. This sampling strategy serves two key purposes, 
By up-weighting $t=0.5$ region, the model prioritizes learning noise correction in the intermediate diffusion stages, where the data exhibits a balance between noise and structure—critical for capturing fine-grained structural details.
Down-weighting $t=0$ and $t=1$ mitigates overfitting to trivial cases (e.g., near-identity mappings) and unstable predictions in highly noisy regimes. 
\fi

\section{ODE Sampler}
We demonstrate the details for the ODE sampler in Alg. \ref{alg:rectflow-sample}.

\begin{algorithm}[H]
\caption{\textbf{ODE Sampling Algorithm}}\label{alg:rectflow-sample}
\textbf{Given:} $\Delta t = \frac{1}{\# \text{Steps}}, \; \sigma_{\text{data}} = 16$, and Diffuser with condition $C$. \\
\textbf{Initial:} $x_0 \sim \mathcal{N}(0, \sigma_{\text{data}} \mathbf{I}), \; t = 0.$
\vspace{4pt}

\While{$t < 1$}{
    $x_t \leftarrow \mathrm{CenterRandomAugmentation}(x_t)$\;
    % $v_t \leftarrow \mathrm{Diffuser}(x_t, t \mid C)$\;
    $x_t^\mathrm{denoised} \leftarrow \mathrm{Diffuser}(x_t, t \mid C)$\;
    $x_t \leftarrow x_t + \Delta t \cdot \mathrm{CalVelocity}(x_t^\mathrm{denoised}, x_t, t)$\;
    % $x_t \leftarrow x_t + \eta \cdot \Delta t \cdot v_t$\;
    $t \leftarrow t + \Delta t$\;
}
\end{algorithm}

\section{Model Configurations}
We list the detailed model architecture and sampling configuration for all of our models in Table \ref{tab:appendix:architecture-config}. 
For the Mini and Tiny models, we apply small architecture configurations with two-step ODE sampler. 
We prune the first (close to the input) 8 Pairformer blocks of Mini, finetune the rest parameters, and get the Tiny model.

\begin{table}[h]
    \centering
    \begin{tabular}{l|c|cc}
    \toprule
    & Protenix & Protenix-Mini & Protenix-Tiny \\
    \hline
    \#MsaBlocks & 4 & 1 & 1 \\
    \#PairformerBlocks & 48 & 16 & 8\\
    \#DiffusionEncoder & 3 & 1 & 1\\
    \#DiffusionTransformer & 24 & 8 & 8 \\
    \#DiffusionDecoder & 3 & 1 & 1 \\
    \hline
    \#Steps & 200 & 2 & 2\\
    $\eta$ & 1.5 & 1 & 1 \\
    $\lambda$ & 1.003 & n/a & n/a \\
    $\gamma_0$ & 0.8 & 0 & 0 \\
    \hline
    TrainingScheme & From Scratch & From Scratch & Finetune \\
    \bottomrule
    \end{tabular}
    \caption{\textbf{The configuration for our models}. We list the key hyper-parameters for the model architecture, diffusion sampling and training.
    }
    \label{tab:appendix:architecture-config}
\end{table}

\section{Confidence Head Results}
In the above sections, we mainly focus on the structure prediction performance and therefore report the median results across 25 samples for each input sequence.
Following the standard confidence head training configurations mentioned in \citep{abramson2024accurate,bytedance2025protenix}, we train the confidence head for our Mini and Tiny models. 
% We describe the selection performance in the following figures. 

\end{document}